\documentclass[10pt,twocolumn,letterpaper]{article}

\usepackage{cvpr}
\usepackage{times}
\usepackage{epsfig}
\usepackage{graphicx}
\usepackage{amsmath}
\usepackage{MyMnSymbol}
\usepackage{amssymb}
\usepackage{color}
\usepackage{paralist}

\usepackage{algorithm2e}
\SetAlFnt{\small}
\SetAlCapFnt{\small}
\SetAlCapNameFnt{\small}
\usepackage{algorithmic}
\algsetup{linenosize=\small}

\usepackage[table,xcdraw]{xcolor}
\usepackage{dsfont}

{

\newcommand{\expo}[1]{\exp \left( #1 \right)}
\newcommand{\E}[1]{\mathbb{E}\left[ #1 \right]}

\newcommand{\thetab}{\boldsymbol{\theta}}

\newcommand{\xb}{\boldsymbol{x}}
\newcommand{\zb}{\boldsymbol{z}}
\newcommand{\yb}{\boldsymbol{y}}
\newcommand{\Hb}{\boldsymbol{H}}
\newcommand{\fb} {\boldsymbol{f}}
\newcommand{\stateb} {S}
\newcommand{\pol}[1]{\pi \left( #1 ;\thetab \right)}

\newcommand{\deriv}[1]{\frac{\partial}{\partial #1}}
\newcommand{\derivT}[2]{\frac{\partial #1}{\partial #2}}

\newcommand{\indi}[1]{\mathds{1}( #1 )}

\usepackage[pagebackref=true,breaklinks=true,letterpaper=true,colorlinks,bookmarks=false]{hyperref}

\cvprfinalcopy 

\ifcvprfinal\pagestyle{empty}\fi
\pagenumbering{gobble}
\begin{document}

\title{PoseAgent: Budget-Constrained 6D Object Pose Estimation \\
via Reinforcement Learning}

\author{Alexander Krull$^1$, Eric Brachmann$^1$, Sebastian Nowozin$^2$,\\
Frank Michel$^1$, Jamie Shotton$^2$, Carsten Rother$^1$\\
$^1$ TU Dresden, $^2$ Microsoft
}

\maketitle

\begin{abstract}
State-of-the-art computer vision algorithms often achieve efficiency by making
discrete choices about which hypotheses to explore next.
This allows allocation of computational resources to promising candidates,
however, such decisions are non-differentiable.  As a result, these algorithms
are hard to train in an end-to-end fashion.
In this work we propose to \emph{learn} an efficient algorithm for the task of 6D object pose estimation. 
Our system optimizes the parameters of an existing state-of-the art pose estimation system using reinforcement learning, where the pose estimation system now becomes the \emph{stochastic policy}, parametrized by a CNN.
Additionally, we present an efficient training algorithm that dramatically reduces computation time.
We show empirically that our learned pose estimation procedure makes better use of limited resources and improves upon the state-of-the-art on a challenging dataset.
Our approach enables differentiable end-to-end training of complex algorithmic pipelines and learns to make optimal use of a given computational budget.
\end{abstract}

\section{Introduction}
Many tasks in computer vision involve \emph{learning a function}, usually
learning to predict a desired output label given an input image.
Advances in deep learning have led to huge progress in solving such tasks.
In particular, convolutional neural networks (CNNs) work well when trained
over large training sets using gradient descent methods to minimize the
expected loss between the predictions and the ground truth labels.

However, important computer vision systems take the form of
\emph{algorithms} rather than being a simple differentiable function:
sliding window search, superpixel partioning, particle filters, and
classification cascades are examples of algorithms realizing complex non-continuous functions.

The \emph{algorithmic approach} is especially useful in situations where
computational budget is limited: an algorithm can dynamically assign its
budget to solving different aspects of the problem, for example, to take
shortcuts in order to spend computation on more promising solutions at the
expense of less promising ones.
We would like to \emph{learn the algorithm}.
Unfortunately, the hard decisions taken in most algorithmic approaches are
non-differentiable, and this means that the structure and parameters of these
efficient algorithms cannot be easily learned from data.

Reinforcement learning (RL)~\cite{sutton1998reinforcement} offers a possible
solution to learning algorithms.
We view the algorithm as the \emph{policy} of an RL agent, \ie a description
of dynamic sequential behaviour.
RL provides a framework to learn the parameters of such behaviour with the
goal of maximizing an expected reward, for example, the accuracy of the
algorithm output.
We apply this perspective on an algorithmic computer vision method.
In particular, we address the problem of 6D object pose estimation and use RL to learn the parameters of a deep algorithmic pipeline to provide the best possible accuracy given a limited computational budget.

\begin{figure*}[t]
\begin{center}
\hspace*{-0.4cm}
  \includegraphics[width=1.0\linewidth]{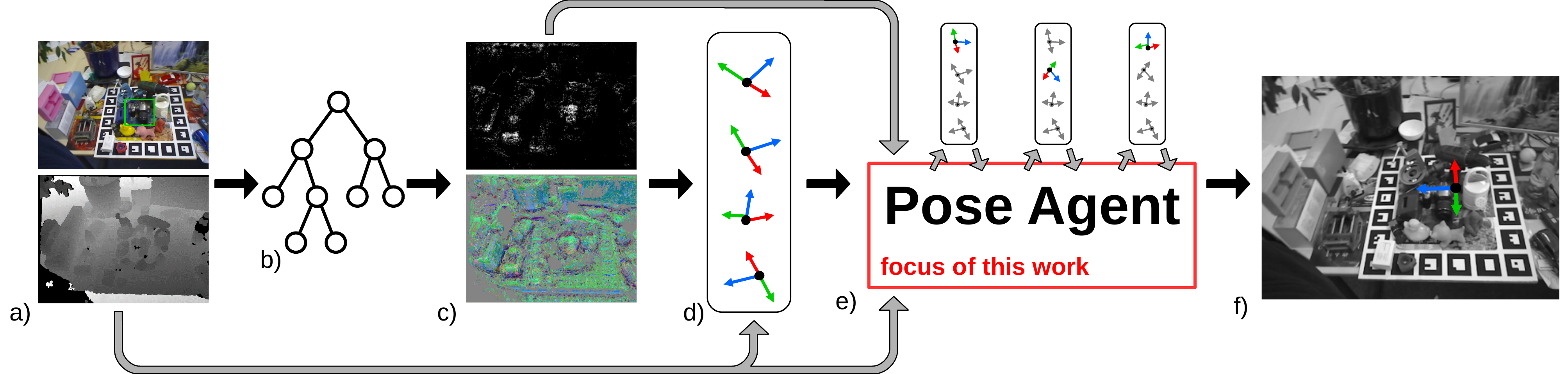}
\end{center}
   \caption{The pose estimation pipeline: a) The input of our system is an RGB-D image. We are interested in the pose of the camera highlighted by the green box.
   b) Similar to \cite{brachmann2014}, the image is processed by a random forest.
   c) The forest outputs dense predictions of object probabilities (top) and object coordinates (bottom). The object coordinates are mapped to the RGB cube for visualization.
   d) We use the predictions together with the depth information to sample a pool of pose hypotheses $\Hb^0$.
   e) An RL agent manipulates the hypotheses pool by repeatedly selecting individual hypotheses to refine. This is the focus of this paper.
   f) The agent outputs a final pose estimate $\tilde{H}$.}
\label{fig:pipeline}
\end{figure*}

\emph{Object pose estimation} is the task of estimating from an image the 3D translation (position) and 3D rotation (orientation) of a specific object relative to its environment.
This task is important in many applications such as robotics and augmented reality where the efficient use of a limited computation budget is an important requirement.
A particular challenge are small, textureless and partially occluded objects in a cluttered environment (see Fig.~\ref{fig:pipeline}).

State-of-the-art pose systems such as the system of
Krull~\etal~\cite{krull2015} generate a pool of pose hypotheses, then score each hypothesis using a pre-trained CNN.
The subset of high-scoring hypotheses get refined and ultimately the
highest-scoring hypothesis is returned as the answer. 
Computationally the refinement step is the most expensive, and there is a
trade-off between the number of refinements allowed and the expected quality
of the result.

Ideally, one would train such state-of-the-art system end-to-end in order to
learn how to use the optimal number of refinements to maximize the expected
success of pose estimation.
Unfortunately, treating the system as a black box with parameters to optimize
is impossible for two reasons:
(i) each selection process is non-differentiable with respect to the scoring function; and
(ii) the loss used to determine whether an estimated pose is correct is also non-differentiable.

We recast pose estimation as an RL problem in order to overcome these difficulties.
We model the pose inference process as an RL agent which we call
\emph{PoseAgent}.
PoseAgent is granted more flexibility than the original system:
it is given a fixed budget of refinement steps, and is allowed to manipulate
its hypothesis pool by selecting individual poses for refinement, until the
budget is spent.
In our PoseAgent model each decision follows a probability distribution over possible actions. This distribution is called the policy and we can differentiate and optimize this continuous policy through the stochastic policy
gradient approach~\cite{sutton1999policygradient}.
As a result of this stochastic approach the final pose estimate becomes a
random variable, and each run of PoseAgent will produce a slightly different
result.

This policy gradient approach is very general and does not require
differentiability of the used loss function.
As a consequence we can directly take the gradient with respect to the
expected loss of interest, \ie the number of correctly estimated poses.
Training in policy gradient methods can be difficult due to the additional
variance of estimated
gradients~\cite{greensmith2004variance,sutton1999policygradient}, because the
additional randomness leads to a bigger variance in the estimated gradients.
To overcome this problem we propose an efficient training algorithm that
radically reduces the variance during training compared to a na\"{\i}ve
technique.

We compare our approach to the state-of-the-art~\cite{krull2015} and achieve
substantial improvements in accuracy, while using the same or smaller average
budget of refinement steps compared to \cite{krull2015}.
In summary our {\bf contributions} are:
\begin{compactitem}
\item To the best of our knowledge, we are the first to apply a policy gradient approach to the object pose estimation problem. 
\item Our approach allows the use of a non-differentiable reward function corresponding to the original evaluation criterion.
\item We present an efficient training algorithm that dramatically reduces the variance during training.
\item We improve significantly upon the best published results on the dataset.
\end{compactitem}

\section{Related Work \label{sec:related_work}}

Below, we first discuss approaches for 6D pose estimation, focusing in particular on object coordinate prediction methods, and then provide a short review of RL methods used in a setting similar to ours.

\subsection{Pose Estimation}
There is a large variety of approaches for 6D object pose estimation.
Traditionally, approaches based on sparse features~\cite{lowe2001local,Martinez_Torres_2010} have been successful, but work well only for textured objects.
Other approaches include template-based methods \cite{hinterstoisser2012accv,rios2013discriminatively}, voting schemes \cite{drost2010model,hinterstoisser2016going}, and CNN-based direct pose regression \cite{Gupta_pose_2015}.

We focus on the line of work called object coordinate regression \cite{brachmann2014}, which provides the basic framework for our approach.
Object coordinate regression was originally proposed for human body pose estimation \cite{taylor2012vitruvian} and camera localization \cite{Shotton_2013}.
In \cite{brachmann2014} a random forest provides a dense pixel-wise prediction for 6D object pose prediction.
At each pixel, the forest predicts whether and where the pixel is located on the surface of the object.
One can then efficiently generate pose hypotheses by sampling a small set of pixels and combining the forest predictions with depth information from an RGB-D camera.

The object coordinate regression methods in \cite{brachmann2014, krull2015, michel2015} score these hypotheses by comparing rendered and observed image patches.
While \cite{brachmann2014, michel2015} use a simple pixel wise distance function, \cite{krull2015} propose a learned comparison: a CNN compares rendered and observed images and outputs an energy value representing a parameter of the posterior distribution in pose space.
Despite their differences in the particular scoring functions, \cite{brachmann2014,michel2015,krull2015} use the same inference technique to arrive at the final pose estimate: they all refine the best hypotheses, re-score them, and output the best one as their final choice.
Our PoseAgent approach can be seen as a generalization of this algorithm, in which the agent selects the hypotheses for refinement repeatedly, each time being able to make a more informed choice.

The work of Krull \etal \cite{krull2015} is the most closely related to our work.
We use a similar CNN construction as Krull \etal, feeding both rendered and observed image patches into to our CNN.
However, we use the output of the CNN as a parameter of the stochastic policy that controls the behaviour of our pose agent.
Moreover, while the training process in \cite{krull2015} is seen as learning the posterior distribution, which is then maximized during testing using the fixed inference procedure, our training process instead modifies the behaviour of the agent directly in order to maximize the number of correctly estimated poses.

\subsection{Reinforcement Learning in Similar Tasks}
RL has traditionally been successful in areas like robotics~\cite{smart2002robots}, control~\cite{abbeel2007helicopter}, advertising, network routing, or playing games.
While the application of RL seems natural for such cases where real agents and environments are involved, RL is increasingly being successfully applied in computer vision systems where the interpretation of the system as an agent interacting with an environment is not always so intuitive.
While we are to our knowledge the first to apply RL for 6D object pose estimation, there are several recent papers that apply RL for 2D object detection and recognition \cite{mnih2014, caicedo2015,mathe2016, ba2015}.

In \cite{mnih2014, caicedo2015}, an agent shifts its area of attention over the image until it makes a final decision.  Instead of moving a single 2D area of attention over search space like \cite{mnih2014, caicedo2015}, we work with a pool of multiple 6D pose hypotheses.
The agent in \cite{mathe2016} focuses its attention by moving a 2D fixation point, though operates on a set of precomputed image regions to gather information and make a final decision.  Our agent instead manipulates its hypothesis pool by refining individual hypotheses.

Caicedo \etal \cite{caicedo2015} use Q-learning, in which the CNN predicts the quality of the available state-action pairs.
Mnih \etal \cite{mnih2014} and Mathe \etal \cite{mathe2016} use a different RL approach based on stochastic policy gradient, in which the behaviour of the agent is directly learned to maximize an expected reward.
We follow \cite{mnih2014, caicedo2015} in using stochastic policy gradient, which allows us to use a non-differentiable reward function, directly corresponding to the final success criterion used during evaluation.

\section{Method \label{sec:methods}}
In this section, we first define the pose estimation task and briefly review the pose estimation pipeline from \cite{brachmann2014, krull2015, brachmann2016}.
We then continue to describe PoseAgent, our reinforcement learning agent, designed to solve the same problem.
Finally, we discuss how to train our agent, introducing our new, efficient training algorithm.

\subsection{Pose Estimation Pipeline \label{sec:pipeline}}
We begin by describing the object pose estimation task.
Given an RGB-D image $\xb$ we are interested in localizing a specific, known, rigid object (Fig.~\ref{fig:pipeline}a). We assume that exactly one object instance is present in the scene. Our goal is to estimate the true pose $H^*$ of the instance, \ie its position in space as well as its orientation. The pose has a total of six degrees of freedom, three for translation and three for rotation. We define the pose as the rigid body transformation that maps a point from the local coordinate system of the object to the coordinate system of the camera.

Our method is based on the work of Krull \etal \cite{krull2015}.
As in \cite{krull2015}, we use an intermediate image representation called object coordinates. By looking at small patches of the RGB-D input image, a random forest (Fig.~\ref{fig:pipeline}b) provides two predictions for every pixel $i$.
Each tree predicts an object probability $p_i \in [0,1]$ as well as a set of object coordinates $\yb_i$ (Fig.~\ref{fig:pipeline}c).
The object probability $p_i$ describes whether the pixel is believed to be part of the object or not. The object coordinates $\yb_i$ represent the predicted position of the pixel on the surface of the object, \ie its 3D coordinates in the local coordinate system of the object.

Again following \cite{krull2015}, we use these forest predictions in a RANSAC-inspired sampling scheme to generate pose hypotheses.
We repeatedly sample three pixels from the image according to the object probabilities $p_i$.
By combining the predicted object coordinate $\yb_i$ with the camera coordinates of the pixels (calculated from the depth channel of the input image), we obtain three 3D-3D correspondences. We calculate a pose hypothesis from these correspondences using the Kabsch algorithm \cite{kabsch1976solution}.
We sample a fixed number $N$ of hypotheses, which are combined in hypothesis pool $\Hb^0=(H^0_1 \dots H^0_N)$ (Fig.~\ref{fig:pipeline}d). The upper index denotes time steps which we will use later in our algorithm.

Krull \etal \cite{krull2015} proposed the following rigid scheme for pose optimization. All hypotheses are scored and the 25 top-scoring hypotheses are refined. Then, the refined hypotheses are scored again, and the best scoring hypothesis is returned as the final pose estimate of the algorithm.

Our paper focuses on improving the process by which the correct pose is found, starting from the same initial hypothesis pool.
We propose to use an RL agent (Fig.~\ref{fig:pipeline}e) to dynamically decide which hypothesis to refine next, in order to make most efficient use of a given computational budget. When its budget is exhausted, the agent selects a final pose estimate (Fig.~\ref{fig:pipeline}f).

\subsection{PoseAgent\label{sec:pose_agent}}

\begin{figure}[t]
\begin{center}
\hspace*{-0.4cm}
  \includegraphics[width=1.1\linewidth]{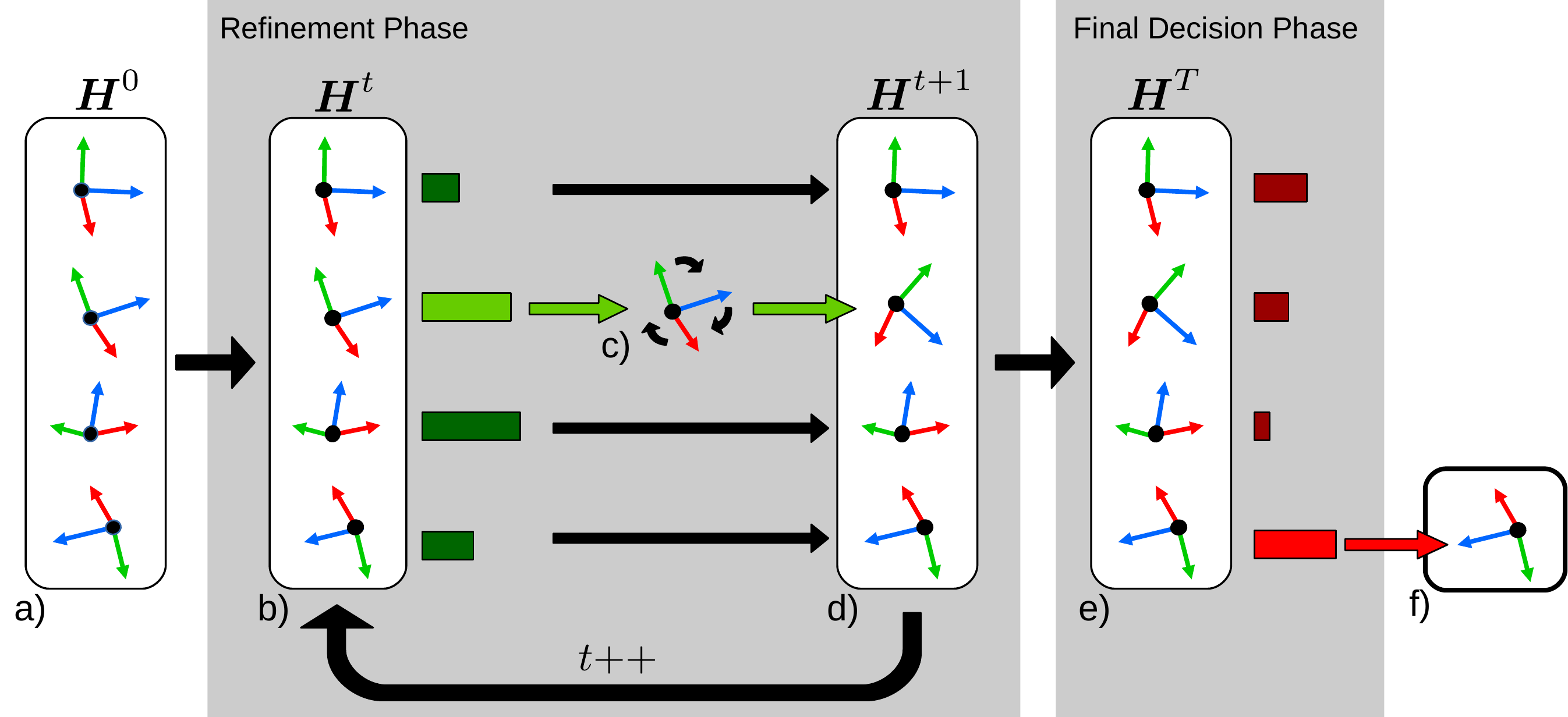}
\end{center}
   \caption{The pose agent inference process:
   a) The initial pool of hypotheses is sampled and handed to the agent.
   b) The agent selects a hypotheses $H^t_{a^t}$ by sampling from the policy $\pol{a^t|\stateb^t}$.
   c) The selected hypothesis is refined.
   d) If refinement budget is left, the refinement phase continues. If the budget is exhausted a final selection is made.
   e) The final selection is made by sampling from the policy $\pol{a^T|\stateb^T}$.
   f) The selected pose $H^T_{a^T}$ is output as final pose estimate.}
\label{fig:poseagent}
\end{figure}

We now describe our RL agent, PoseAgent, and how it performs inference. 
An overview of the process can be found in Fig.~\ref{fig:poseagent}.
The agent operates in two phases: (i) the refinement phase, in which the agent chooses individual hypotheses to undergo the expensive refinement step; and (ii) the final decision phase, in which it has to decide which pose should be selected as final output. In the following, we discuss both phases in detail.

Inference begins with the {\bf refinement phase}.
The pose agent starts with a pool $\Hb^0=(H^0_1 \dots H^0_N)$ of hypotheses which have been generated as described in Sec.~\ref{sec:pipeline}, and a fixed budget $B^0$ of possible refinement steps. 

At each time step $t$, the agent chooses one hypothesis index $a^t$, which we call an action. The chosen hypothesis is refined and the next time step begins. We limit the maximum number of times the agent may choose the same hypothesis for refinement to $\tau_{\max}$. Hence, over time, the pool of actions (resp. hypotheses) the agent may choose for refinement decreases. We denote the set of possible actions $A^t=\{ a \in \{ 1, \dots ,N \}| \tau_a^t < \tau_{\max} \}$, where $\tau_a^t$ denotes how many times hypothesis $a$ has been refined before time $t$. Subsequently, the agent modifies the hypothesis pool by refining  hypothesis $H^{t+1}_{a^t} = g(H^t_{a^t})$, where $g(\cdot)$ is the refinement function.
All other hypotheses remain unchanged $H^{t+1}_a = H^{t}_a ~\forall a \neq a^t$.

We perform refinement as follows (see also \cite{krull2015}).
We render the object in pose $H^t_{a^t}$.
Each pixel within the rendered mask is tested for being an inlier.\footnote{A pixel $i$ is tested for being an inlier for pose $H$ by transforming its predicted object coordinates $\yb_i$ to camera space using $H$. If the Euclidean distance between resulting camera coordinates and the observed coordinates at the pixel is below a threshold the pixel is considered an inlier.}
All inlier pixels are used to re-calculate the pose with the Kabsch algorithm. We repeat this procedure multiple times for the single, chosen hypothesis until the number of inlier pixels stops increasing or until the number $m^t$ of executed refinement steps exceeds a maximum $m_\text{max}$.
The budget is decreased by the number of refinement steps performed, $B^{t+1}=B^t - m^t$.
The agent proceeds choosing refinement actions until $B^t<m_\text{max}$, in which case further refinement may exceed the total budget $B^0$ of refinement steps. 

When this point has been reached, the refinement phase terminates, and the agent enters the {\bf final decision phase} in which the agent chooses a hypothesis as the final output. We denote the final action as $a^T \in \{ 1 \dots N \}$ and the final pose estimate as $\tilde{H}=H^T_{a^T}$.
The agent receives a reward of $r=1$ in case the pose is correct or a negative reward of $r=-1$ otherwise. We use the pose correctness criterion from \cite{hinterstoisser2012accv}.

In the following, we describe how the agent makes its decisions.
During both, the refinement phase and the final decision phase, the agent chooses from the hypothesis pool.
We describe the agent behaviour by a probability distribution $\pol{a^t|\stateb^t}$ referred to as "policy". Given the current state $\stateb^t$, which contains information about the hypothesis pool and the input image $\xb$, the agent selects a hypothesis by drawing a sample from the policy. The vector $\thetab$ of learnable parameters consists of CNN weights (described in Sec.~\ref{sec:cnn}). We will first give details on the state space $\stateb^t$ before explaining policy $\pol{a^t|\stateb^t}$.

\subsubsection{State Space}

We model our state space in a way that allows us to use our new, efficient training algorithm, described in Sec.~\ref{sec:eff:train}.
We assume that the current state $\stateb^t$ of the hypothesis pool decomposes as $\stateb^t=(s^t_1, \dots s^t_N)$, where $s^t_a$ will be called the state of hypothesis $H^t_a$.
The state of an hypothesis contains the original input image $\xb$, the forest prediction $\zb$ for the image, the pose hypothesis $H^t_a$, as well as a vector $\fb^t_a$ of additional context features of the hypothesis (see Sec.~\ref{sec:cnn}). In summary, this gives $s^t_a=(\xb,\zb, H^t_a, \fb^t_a)$.

\subsubsection{Policy}
Our agent makes its decisions using a softmax policy.
The probability of choosing a particular action $a^t$ during the refinement phase is given by
\begin{equation}
	\label{eq:softmax}
	\pol{a^t|\stateb^t}= \frac{\expo{E(s_{a^t}^t;\thetab)}}{\sum_{a \in A^t} \expo{E(s_{a}^t;\thetab)}},
\end{equation}
where $E(s_{a}^t;\thetab)$ will be called the energy of the state $s_a^t$.
We will abbreviate it as $E_a^t=E(s_{a}^t;\thetab)$.
The energy of a state in the softmax policy is a measure of how desirable it is for the agent to refine the hypothesis.
We use the same policy in the final decision phase, but with a different energy ${E'}(s_{a}^t;\thetab)$ abbreviated by ${E'}_a^t$. We use a CNN to predict both energies, $E_a^t$ and ${E'}_a^t$. In the next section, we discuss the CNN architecture and how it governs the behaviour of the agent.

\subsubsection{CNN Architecture \label{sec:cnn}}

\begin{figure}[t]
\begin{center}
\hspace*{-0.5cm}
  \includegraphics[width=1.1\linewidth]{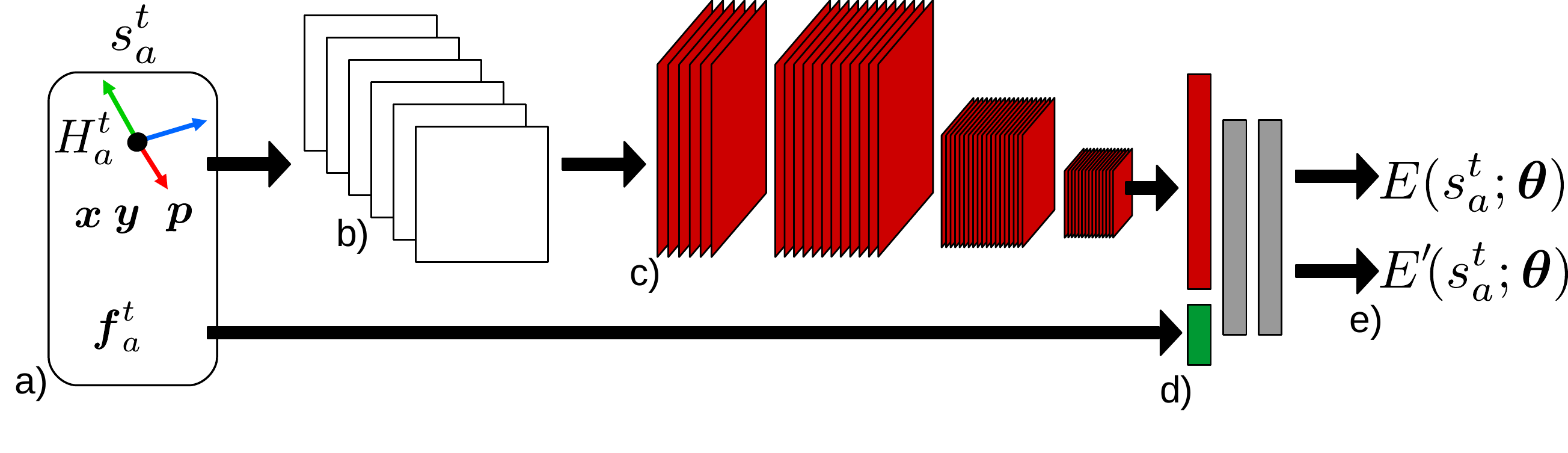}
\end{center}
   \caption{The CNN architecture: a) The system takes a pose hypothesis $H_a^t$ and the additional features $\fb^t_a$ encoding the context and history of the pose as input. b) We use the hypothesis to render the object and to cut out patches in the observed images. c) The images are processed by multiple convolutional layers. d) We concatinate the output of the convolutional layers with the features $\fb^t_a$. The result is fed into multiple fully-connected layers. e) The network predicts two energy values: $E(s_a^t)$ to be used in the refinement phase and $E'(s_a^t)$ to be used in the final decision.}
\label{fig:inference}
\end{figure}

We give an overview of the CNN architecture used in this work in Fig.~\ref{fig:inference}.
As in \cite{krull2015}, the CNN compares rendered and observed images.
We use the same six input channels as in \cite{krull2015}, namely: the rendered depth channel, the observed depth channel, a rendered segmentation channel, the object probability channel, a depth mask, and a single channel holding the difference between object coordinates.

There are however two major differences in our CNN compared to the one used in \cite{krull2015}.
Firstly, while Krull \etal predict a single energy value of a pose, we jointly predict two separate energy values: one energy $E^t_a$ for the refinement phase and one energy $E'^t_a$ for the final decision phase.

Secondly, we input additional features to the network by concatenating them to the first fully connected layer.
The features are:
The number of times the hypothesis has already been selected for refinement,
The distance the hypothesis has moved during its last refinement and
the average distance of the hypothesis before refinement to all other hypotheses in the original pool.  

Our CNN consists of the following layers: 128 kernels of size $6\times 3 \times 3$, 256 kernels of size $128\times 3 \times 3$, a $2 \times 2$ \emph{max-pooling} layer, 512 kernels of size $256 \times 3 \times 3$, a max-pooling operation over the remaining size of the image, finally 3 fully connected layers.
The features $\fb^t_a$ are concatenated to the first fully connected layer, as shown in Fig.~\ref{fig:inference}.
Each layer, except the last is followed by a \emph{tanh} operation.

\subsection{Policy Gradient Training}
We will now discuss the training procedure for our PoseAgent. First, we will give a general introduction of policy gradient training, and then apply the approach to the PoseAgent.
Finally, we will introduce an efficient algorithm that greatly reduces variance during training and makes training feasible.

The goal of the training is the maximization of the expected reward $\E{r}$.
This expected value depends on the environment as well as on the policy of our agent.
In stochastic policy gradient methods one attempts to approximate the gradient with respect to the policy parameters $\thetab$.
Note that since we are dealing with the expected value it becomes possible calculate derivatives, even if the reward function itself is non differentiable. 
By making use of the equality $\deriv{\theta_j}p(x;\theta_j)=p(x;\theta_j)\deriv{\theta_j}\ln p(x;\theta_j)$, we can write the derivative of the expected reward with respect to each parameter $\theta_j$ in $\thetab$ as
\begin{equation}
		 \deriv{\theta_j} \E {r}  = \E { r \deriv{\theta_j} \ln p(s^{1:T},a^{1:T};\thetab)},
		 \label{eq:expected_deriv_full}
\end{equation}
where $p(s^{1:T},a^{1:T};\thetab)$ is the probability of a particular sequence of states $s^{1:T}=(s^1 \dots s^T)$ and actions $a^{1:T}=(a^1 \dots a^T)$ to occur.

Because of the Markov property of the environment, it is possible to decompose the probability and rewrite it as
\begin{equation}
	\begin{split}
 	    \deriv{\theta_j} \E {r}  =
 	     	& \E { r \sum_{t=0}^T  \deriv{\theta_j} \ln \pol{a^t|\stateb^t} }
	\end{split}. \label{eq:expected_deriv}
\end{equation}
Following the REINFORCE algorithm \cite{williams1992}, we approximate Eq.~\ref{eq:expected_deriv} using sampled sequences $(\stateb^{1:T_k}_k,a^{1:T_k}_k)$, generated by running the agent, as described in Sec.~\ref{sec:pose_agent}, on training images,
\begin{equation}
	\begin{split}
 			\deriv{\theta_j} \E {r} \approx \frac{1}{M} \sum_{k=1}^M r_k \sum_{t=0}^{T_k}  \deriv{\theta_j} \ln \pol{a^t_k|\stateb^t_k} 
	\end{split}, \label{eq:sampled_deriv}
\end{equation}
where $T_k$ is the number of steps in the sequence and $r_k$ is the reward achieved in the sequence.

\subsubsection{Efficient Gradient Calculation}
\label{sec:eff:train}
We will now introduce an algorithm (Alg.~\ref{alg:training}), to dramatically reduce the variance of estimated  gradients, by allowing us to use a higher number of sequences $M$, in a given time.
The basic idea is to make use of the special decomposable structure of the state space and our policy.
The advantage of our algorithm compared to the na\"ive implementation is illustrated in Fig.~\ref{fig:variance}.

Starting from a hypothesis pool $\Hb^0=(H_1^0 \dots H_N^0)$, only a finite number of different hypothesis states \mbox{$s^{\tau}_a| a \in \{ 1, \dots N \}, \tau \in \{ 0, \dots \tau_{\max} \}$} can occur during a run of our PoseAgent.
Here, $s^{\tau}_a= (H^{\tau}_a, \fb^{\tau}_a)$ shall denote the state of hypothesis $a$ after it has been refined $\tau$ times.

The algorithm pre-computes all possibly occurring hypothesis states $s^{\tau}_a$, and predicts all corresponding energy values $E^{\tau}_a$\footnote{
To improve readability, we will not differentiate between $E^{\tau}_a$ and $E'^{\tau}_a$ in this section.}
in advance using the CNN.

While this comes with some computational expense, it allows us to rapidly sample large numbers of sequences without having to re-evaluate the energy function.

To illustrate why this is possible, let us now reconsider the calculation of the derivatives in Eq.~\ref{eq:sampled_deriv}. Using the chain rule, we can write them as
\begin{equation}
	\deriv{\theta_j} \ln \pol{a^t|\stateb^t}
	=\sum_{a \in A^t} \derivT{E^t_{a}}{\theta_j} \deriv{E^t_{a}} \ln \pol{a^t|\stateb^t},
	\label{eq:policy_deriv_alt}
\end{equation}
where
\begin{equation}
   \deriv{E^t_{a}} \ln \pol{a^t|\stateb^t} =
   \begin{cases}
     1-\pol{a^t|\stateb^t}  & \text{if } a=a^t\\
      -\pol{a^t|\stateb^t}  & \text{else} \\
   \end{cases}.
   \label{eq:select}
\end{equation}

We can now rearrange Eq.~\ref{eq:sampled_deriv} as sum over possible hypothesis states
\begin{equation}
		 \sum_{\tau=0}^{\tau_{\max}} \sum_{a=1}^N \derivT{E^\tau_{a}}{\theta_j}  
		 	\frac{1}{M}\underbrace{\sum_{k=1}^M \sum_{t=1}^{T_k} \indi{\tau^t_{a,k}=\tau}  \deriv{E^\tau_{a}} \ln \pol{a^t_k|\stateb_k^t} r_k}_{D(a,\tau)}
	. \label{eq:sampled_deriv_2}
\end{equation}
Here, $\indi{\tau^t_{a,k}=\tau}$ is the indicator function. It has the value $1$ only when the hypothesis $a$ at time $t$ in sequence $k$ has been selected for refinement exactly $\tau^t_{a,k}=\tau$ times. It has the value $0$ in any other case.

Our algorithm works by first calculating the inner sums in Eq.~\ref{eq:sampled_deriv_2} and storing the results in the entries $D(a,\tau)$ of a table $D$.
We compute these sums with a single iteration over all sequences $k$ and all time steps $t$.
The accumulation of these values is computationally cheap, because it does not not require any rendering or involvement of the CNN.

This structure allows us to increase the number of sampled sequences $M$ without much computational cost.
The algorithm can process an arbitrary amount of sequences using only a single back propagation pass of the CNN for each possible hypothesis state $s^t_a$.
In a na\"ive implementation, the number of required forward-backward passes would increase linearly with the number of sampled sequences.

Let us look at the algorithm in detail.
It consists of three parts:

\noindent {\bf Initialization Phase:} We generate the original hypothesis pool as described in \ref{sec:pipeline}.
Then, we refine all hypotheses $\tau_{\max}$ times and predict the energy values $E^{\tau}_a$ for all of them using the CNN.

\noindent {\bf Sampling Phase:} We sample sequences $(s^{1:T}_k,a^{1:T}_k)$ as described in Sec.~\ref{sec:pose_agent}, using the precomputed energies.
We observe the reward $r_k$ for each sequence.
Then, we calculate for each time $t$, each selected hypothesis $a^t_k$ and each possible hypothesis $a$ the derivative $\deriv{E^\tau_{a}} \ln \pol{a^t_k|\stateb_k^t} r_k$ using Eq.~\ref{eq:select}.
We accumulate the results in the corresponding table entries $D(a,\tau^t_{a,k})$.
This corresponds to the inner sums in Eq.~\ref{eq:sampled_deriv_2}.

\noindent{\bf Gradient Update Phase:} We once more process each of the hypothesis states $s^{\tau}_a$ with the CNN and use standard back propagation to calculate $\derivT{E^\tau_{a}}{\theta_j}$.
We multiply the results with $D(a,\tau)$ and accumulate them up in another table $G$ to obtain the final gradients.
This corresponds to the outer sums in Eq.~\ref{eq:sampled_deriv_2}.

\begin{algorithm}
\algsetup{linenosize=\tiny}
\label{alg:training}
\SetKwBlock{Initial}{Initialization Phase:}{}
\SetKwBlock{Sampling}{Sampling Phase:}{}
\SetKwBlock{Grads}{Gradient Update Phase:}{}

\Initial{
 generate hypothesis pool $\Hb^0$\; 
 refine each hypothesis {${\tau}_{\max}$ times}\;
 calculate and store $E^{\tau}_a$\;
 initialize table entries $D(a,\tau)\gets0$ and $G(j)\gets0$
 
}

\Sampling{
 \For{$k=1:M$}{
 	sample path $(s_{k}^{1:T_k}, a_{k}^{1:T_k})$ using $E^{\tau}_a$ \;
 	receive reward $r_k$\; 
 	\For{$t=1:T_k, a=1:N$}{
 			\mbox{$D(a,\tau_{a,k}) \small{\gets} D(a,\tau_{a,k}) +\deriv{E^\tau_{a}} \ln \pol{a^t_k|\stateb_k^t} r_k$}
 	}
 }
}

\Grads{
\For{$\tau=0:\tau_{\max}$~,~ $a=1:N$}{
	calculate $\derivT{E^\tau_{a}}{\theta_j}$ via back propagation\;
	\For{\emph{all CNN parameters} $j$}{
	 $G(j) \gets G(j) + \derivT{E^\tau_{a}}{\theta_j} \frac{1}{m} D(a,\tau)$;
	 }
}
{\bf Output:} $G(j)\approx \deriv{\theta_j} \E {r}$;
}
 \caption{Efficient Gradient Calculation}
\end{algorithm}

\section{Experiments \label{sec:experiments}}
In the following we will describe the experiments to compare our method to the baseline system from \cite{krull2015}.
Our experiments confirm, that our learned inference procedure is able to use its budget in a more efficient way.
It outperforms the \cite{krull2015}, while using on average a smaller number refinement steps. 

Additionally we will describe an experiment regarding the efficiency of our training algorithm compared to a na\"ive implementation of the REINFORCE algorithm.
We find, that our algorithm can dramatically reduce the gradient variance during training.

We conducted our experiments on the dataset introduced in \cite{krull2014}.
It features six RGB-D sequences of hand held sometimes strongly occluded objects.

\subsection{Training and Validation Procedure}
We train our system on the \emph{samurai 1} sequence and, as \cite{krull2015} omit the first 400 frames to achieve a higher percentage of occluded images.

We train our system with two different parameter settings:
Using a hypothesis pool size of $N=210$, which is the setting used in \cite{krull2015}, and a larger pool size of $N=420$.

To determine the adequate size of the budget $B^0$ of refinement steps, we ran the system from \cite{krull2015} on our validation set and determined the average number of refinement steps it used. 
We set our budget during training to the resulting number $B^0=77$.

During training we allow a hypothesis to be chosen $\tau_{\max}=3$ times for refinement.
We set the maximum number of refinement steps per iteration to $m_{\max}=10$.

Using stochastic gradient descent, we go through our training images in random order and run Algorithm~\ref{alg:training} to approximate the gradient.
We sample $M=50k$ sequences for every image.
An additional $50k$ sequences are used to estimates the average reward for the image, which is then subtracted from the reward to further reduce variance \cite{sutton1999policygradient,mnih2014}.
We perform a parameter update after every image.
Starting with an initial learning rate of $\lambda^0=25\cdot 10^{-4}$, we reduce it according to $\lambda^l=\lambda^0 / (1+l\nu)$, with $\nu=0.01$.
We use a fixed momentum of $0.9$.

We skip images in which none of the hypotheses from the pool lead to a correct pose after being refined $\tau_{\max}$ times, and in which more than 10\% of the hypotheses from the pool lead to a correct pose.
Such images contribute little, because they are impossibly or to easily solved.

We run the training procedure for 96 hours on an Intel E5-2450 2.10GHz with Nvidia Tesla K20x GPU and save a snapshot every 50 training images.
To avoid over-fitting, we test these saved snapshots on our validation set and choose the model with the highest accuracy.
In order to reduce the computational time during validation, we considered only images in which the object was at least 5\% occluded\footnote{according to the definition from \cite{krull2015}}.

\subsection{Additional Baselines}
Two demonstrate the advantage of dynamically distributing a given computational budget, we implemented two cut-down versions of PoseAgent, which serve as additional baselines.  
The baseline method abbreviated as \emph{RandRef} randomly selects
a hypothesis to refine at every iteration. When the budget
is exhausted, it chooses the hypothesis with the best predicted final selection energy $E'(s^t_a; \thetab)$.

The baseline \emph{BestRef} directly picks the hypothesis with the best $E(s^t_a; \thetab)$, refines it until the budget is exhausted and outputs it as final decision. We used the best performing settings when running the baselines: ($\tau_{max}$ = 6, $m_{max}$ = 5) for pool size N = 210 and ($\tau_{max}$ = 7, $m_{max}$ = 4) for pool size
N = 420.

\subsection{Testing Conditions}

We compared both versions of our model, using $N=210$ and $N=420$, against the system of \cite{krull2015} using the corresponding pool size.
In all experiments with the baseline method \cite{krull2015}, we use the identical CNN with the original weights trained by Krull \emph{et al.} in \cite{krull2015} on the \emph{samurai 2} sequence.
This is the network that \cite{krull2015} reports the best results for.

Apart from the pool size, we used the identical testing conditions as in \cite{krull2015}, including the same random forest originally trained in \cite{brachmann2014}.
To classify a pose in correct or false we use the same point-distance-based criterion used in \cite{krull2015}. 
A pose is considered correct, when the average distance between the vertices of the 3D model in the ground truth pose and the evaluated pose is below a threshold.

While the number of refinement steps in our setting is restricted, the method of \cite{krull2015} does not provide any guarantees on how many refinement steps are used.  
To ensure a fair comparison, we first ran the method from \cite{krull2015} and recorded the average number of refinement steps that it required on each test sequence.
When running our method, we set the budget for each sequence to this recorded value, making sure that PoseAgent could never use more refinement steps than \cite{krull2015}.
The total average number of refinement steps required by both methods can be found Tabs.~\ref{tab:210} and \ref{tab:420}.

We evaluate our method using different parameters for $\tau_{\max}$ and $m_{\max}$, so that $\tau_{\max} \cdot m_{\max} \approx 30$. Meaning that a each pose can have an approximate maximum of $30$ refinement steps.
A higher value of $\tau_{\max}$ (and lower value of $m_{\max}$) means that PoseAgent can make more fine grained decisions on where to spend its budget. 
We use the following combinations for the two values ($\tau_{\max}$=3,$m_{\max}=10),$ $(\tau_{\max}=5,m_{\max}=6),$ $(\tau_{\max}=6,m_{\max}=5),$ $(\tau_{\max}=7,m_{\max}=4)$.

\subsection{Results}
The results of our experiments can be seen in Tabs.~\ref{tab:210} and \ref{tab:420}. 
PoseAgent is able to improve the best published results on the dataset by a total of {\bf 10.56\%} (comparing 60.06\% from Tab.~\ref{tab:210} with 70.62\% from Tab.~\ref{tab:420}).
When we compare our method to \cite{krull2015} working on the same hypothesis pool size we are still able to outperform it. 
With the original pool size of $N=210$ by {\bf 2.12\%} and the increased pool size of $N=420$ by {\bf 2.59\%}.

Note, that the budget is set in a way, that is extremely restrictive, ensuring that PoseAgent can never use more refinement steps than \cite{krull2015} uses on average.

In both settings ($N=210$ and $N=420$) there appears to be a trend, that an increase of $\tau_{\max}$, which corresponds to a more fine grained control of PoseAgent, leads to an improvement in accuracy.
The only exception here is $\tau_{\max}=7$ in the $N=210$ setting.
It should be noted that PoseAgent was trained with a different setting of $\tau_{\max}=3$ and was able to generalize to the different settings used during testing.

We measured the average run time of the method (using CPU rendering) to be between 17 (Samurai 2) and 34 (Cat 2) seconds per image on an Intel E5-2450 2.10GHz with NVidia Tesla K20x GPU using a hypothesis pool of $N=420$.

\begin{table}[]
\centering
\resizebox{8.8cm}{!}{%
\begin{tabular}{llllllll}
\cline{3-6}
                                      & \multicolumn{1}{l|}{}                                      & \multicolumn{4}{c|}{Ours}                                                                                                                                                                                                       &                              &                              \\ \cline{2-8} 
\multicolumn{1}{l|}{}                 & \multicolumn{1}{l|}{Krull \emph{et al.}}                                 & \multicolumn{1}{l|}{$\tau_{\max}$=3}              & \multicolumn{1}{l|}{$\tau_{\max}$=5}                       & \multicolumn{1}{l|}{$\tau_{\max}$=6}                       & \multicolumn{1}{l|}{$\tau_{\max}$=7}              & \multicolumn{1}{l|}{RandRef} & \multicolumn{1}{l|}{BestRef} \\ \hline
\multicolumn{1}{|l|}{Cat 2}           & \multicolumn{1}{l|}{{\color[HTML]{000000} 63.05}}          & \multicolumn{1}{l|}{{\color[HTML]{000000} 67.81}} & \multicolumn{1}{l|}{{\color[HTML]{000000} 68.61}}          & \multicolumn{1}{l|}{{\color[HTML]{000000} \textbf{71.52}}} & \multicolumn{1}{l|}{{\color[HTML]{000000} 68.74}} & \multicolumn{1}{l|}{45.17}   & \multicolumn{1}{l|}{49.27}        \\ \hline
\multicolumn{1}{|l|}{Samurai 2}       & \multicolumn{1}{l|}{{\color[HTML]{000000} \textbf{60.30}}} & \multicolumn{1}{l|}{{\color[HTML]{000000} 51.66}} & \multicolumn{1}{l|}{{\color[HTML]{000000} 53.32}}          & \multicolumn{1}{l|}{{\color[HTML]{000000} 54.82}}          & \multicolumn{1}{l|}{{\color[HTML]{000000} 51.66}} & \multicolumn{1}{l|}{31.73}        & \multicolumn{1}{l|}{34.88}        \\ \hline
\multicolumn{1}{|l|}{Toolbox 2}       & \multicolumn{1}{l|}{{\color[HTML]{000000} 52.96}}          & \multicolumn{1}{l|}{{\color[HTML]{000000} 52.07}} & \multicolumn{1}{l|}{{\color[HTML]{000000} \textbf{60.06}}} & \multicolumn{1}{l|}{{\color[HTML]{000000} 54.44}}          & \multicolumn{1}{l|}{{\color[HTML]{000000} 59.76}} & \multicolumn{1}{l|}{35.50}   & \multicolumn{1}{l|}{32.84}   \\ \hline
\multicolumn{1}{|l|}{Total}           & \multicolumn{1}{l|}{{\color[HTML]{000000} 60.06}}          & \multicolumn{1}{l|}{{\color[HTML]{000000} 58.94}} & \multicolumn{1}{l|}{{\color[HTML]{000000} 61.47}}          & \multicolumn{1}{l|}{{\color[HTML]{000000} \textbf{62.18}}} & \multicolumn{1}{l|}{{\color[HTML]{000000} 60.88}} & \multicolumn{1}{l|}{38.47}        & \multicolumn{1}{l|}{40.88}        \\ \hline
                                      & {\color[HTML]{000000} }                                    & {\color[HTML]{000000} }                           & {\color[HTML]{000000} }                                    & {\color[HTML]{000000} }                                    & {\color[HTML]{000000} }                           &                              &                              \\ \cline{1-6}
\multicolumn{1}{|l|}{Avg. ref. steps} & \multicolumn{1}{l|}{{\color[HTML]{000000} 68.71}}          & \multicolumn{1}{l|}{{\color[HTML]{000000} 62.94}} & \multicolumn{1}{l|}{{\color[HTML]{000000} 65.91}}          & \multicolumn{1}{l|}{{\color[HTML]{000000} 66.60}}          & \multicolumn{1}{l|}{{\color[HTML]{000000} 67.20}} &                              &                              \\ \cline{1-6}
\end{tabular}
}
\vspace{0.15cm}
\caption{Percent correct poses using a hypothesis pool of $N=210$\label{tab:210}}
\end{table}

\begin{table}[]
\centering
\resizebox{8.8cm}{!}{%
\begin{tabular}{llllllll}
\cline{3-6}
                                      & \multicolumn{1}{l|}{}               & \multicolumn{4}{c|}{Ours}                                                                                                                                                                                          &                              &                              \\ \cline{2-8} 
\multicolumn{1}{l|}{}                 & \multicolumn{1}{l|}{Krull \emph{et al.}}          & \multicolumn{1}{l|}{$\tau_{\max}$=3} & \multicolumn{1}{l|}{$\tau_{\max}$=5}              & \multicolumn{1}{l|}{$\tau_{\max}$=6}                       & \multicolumn{1}{l|}{$\tau_{\max}$=7}                       & \multicolumn{1}{l|}{RandRef} & \multicolumn{1}{l|}{BestRef} \\ \hline
\multicolumn{1}{|l|}{Cat 2}           & \multicolumn{1}{l|}{72.98}          & \multicolumn{1}{l|}{74.70}           & \multicolumn{1}{l|}{{\color[HTML]{000000} 74.97}} & \multicolumn{1}{l|}{{\color[HTML]{000000} 76.29}}          & \multicolumn{1}{l|}{{\color[HTML]{000000} \textbf{78.01}}} & \multicolumn{1}{l|}{54.17}   & \multicolumn{1}{l|}{56.03}        \\ \hline
\multicolumn{1}{|l|}{Samurai 2}       & \multicolumn{1}{l|}{\textbf{66.45}} & \multicolumn{1}{l|}{59.30}           & \multicolumn{1}{l|}{{\color[HTML]{000000} 59.63}} & \multicolumn{1}{l|}{{\color[HTML]{000000} 58.80}}          & \multicolumn{1}{l|}{{\color[HTML]{000000} 61.13}}          & \multicolumn{1}{l|}{39.87}        & \multicolumn{1}{l|}{40.86}        \\ \hline
\multicolumn{1}{|l|}{Toolbox 2}       & \multicolumn{1}{l|}{64.79}          & \multicolumn{1}{l|}{65.38}           & \multicolumn{1}{l|}{{\color[HTML]{000000} 68.93}} & \multicolumn{1}{l|}{{\color[HTML]{000000} \textbf{71.60}}} & \multicolumn{1}{l|}{{\color[HTML]{000000} 71.01}}          & \multicolumn{1}{l|}{52.07}   & \multicolumn{1}{l|}{53.85}   \\ \hline
\multicolumn{1}{|l|}{Total}           & \multicolumn{1}{l|}{68.03}          & \multicolumn{1}{l|}{67.37}           & \multicolumn{1}{l|}{{\color[HTML]{000000} 68.32}} & \multicolumn{1}{l|}{{\color[HTML]{000000} 69.14}}          & \multicolumn{1}{l|}{{\color[HTML]{000000} \textbf{70.62}}} & \multicolumn{1}{l|}{48.67}        & \multicolumn{1}{l|}{50.21}        \\ \hline
                                      &                                     &                                      & {\color[HTML]{000000} }                           & {\color[HTML]{000000} }                                    & {\color[HTML]{000000} }                                    &                              &                              \\ \cline{1-6}
\multicolumn{1}{|l|}{Avg. ref. steps} & \multicolumn{1}{l|}{71.12}          & \multicolumn{1}{l|}{65.00}           & \multicolumn{1}{l|}{{\color[HTML]{000000} 68.04}} & \multicolumn{1}{l|}{{\color[HTML]{000000} 68.66}}          & \multicolumn{1}{l|}{{\color[HTML]{000000} 69.39}}          &                              &                              \\ \cline{1-6}
\end{tabular}
}
\vspace{0.15cm}
\caption{Percent correct poses using a hypothesis pool of $N=420$
\label{tab:420}}
\end{table}

\subsection{Efficiency of the Training Algorithm}
In order to investigate the efficiency of out training algorithm compared to a na\"ive implementation of the REINFORCE algorithm, we conducted the following experiment:

We ran our training algorithm as well as the na\"ive implementation up to 100 times on a single training image without updating the network.

To estimate the variance of the gradient, we calculated the standard deviation of 1000 randomly selected elements from the resulting gradient vector of the CNN and averaged them.
We recorded the required computation time to process the image on an Intel E5-2450 2.10GHz with Nvidia Tesla K20x GPU.

\begin{figure}[t]
\begin{center}
     \includegraphics[width=0.7\linewidth]{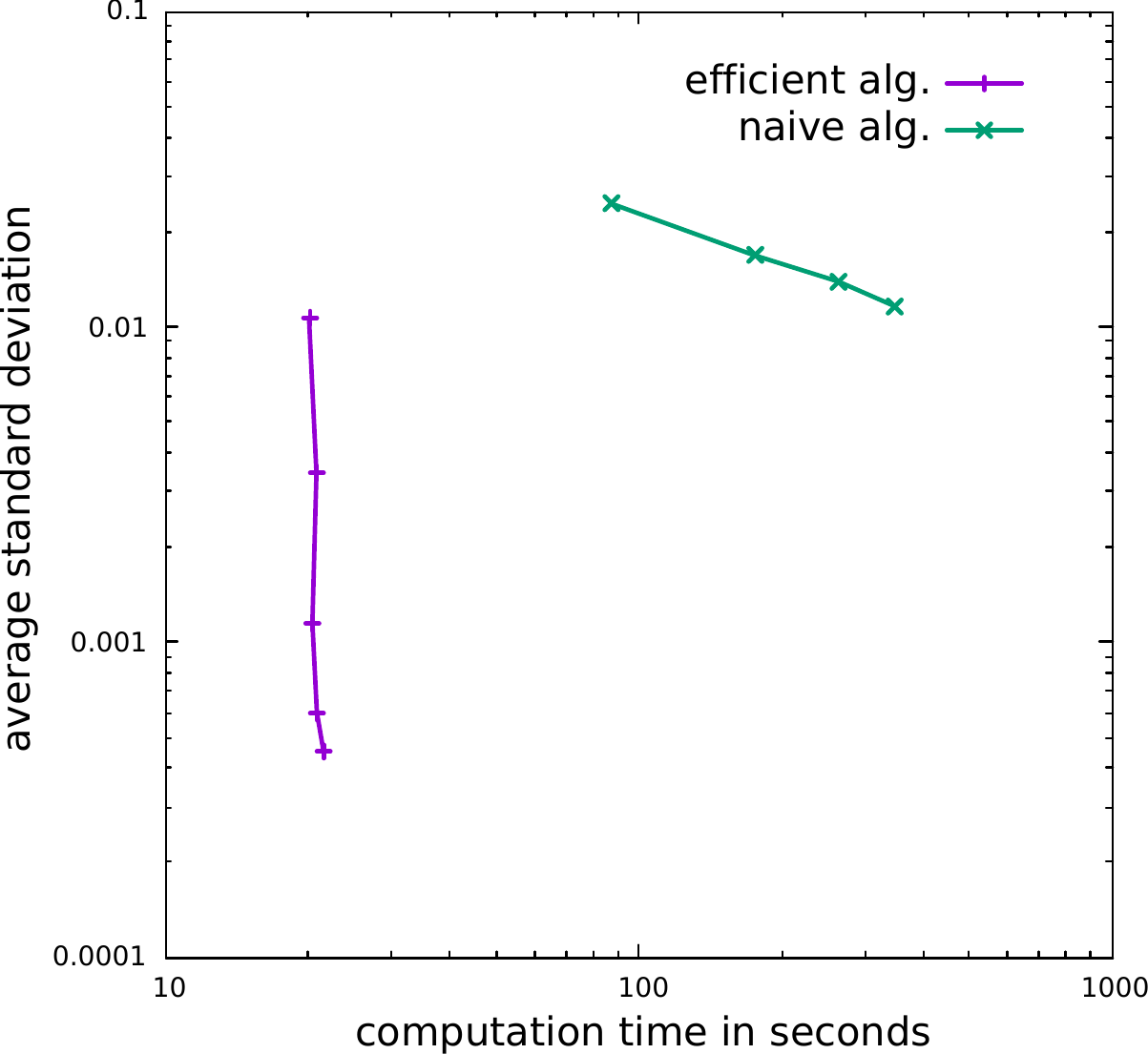}
\end{center}
   \caption{Observed gradient variance during training as a function of time: Our method is able process dramatically more sequences with almost no increase in computation time compared to a na\"ive implementation of the REINFORCE algorithm.
   The result is a drastically reduced gradient variance}
\label{fig:variance}
\end{figure}

The process was repeated for $M=5$, $M=50$, $M=500$, $M=5000$ and $M=50000$ sequences in case of the efficient algorithm.
In case of the na\"ive implementation we used $M=1$, $M=2$, $M=3$ and $M=4$ sequences. 
To keep the computation time in reasonable limits we used a reduced setting with a hypothesis pool size of $N=21$ for both methods.
As can be seen in Fig.~\ref{fig:variance} our algorithm allows us to reduce variance greatly with almost no increase in computation time.
 
\section{Conclusion \label{sec:conclusion}}
We have demonstrated a method learn the algorithmic inference procedure in a pose estimation system using a policy gradient method.
Our system learns to make efficient use of a given budget and is able to outperform the original system, while using on average less computational resources.
We have presented an efficient algorithm for the gradient approximation during training. The algorithm is able to sharply reduce  gradient variance, without a significant increase in computation time.

We see multiple interesting future directions of research in the context of our system. 
(i) One could investigate a soft version of PoseAgent, which is not working with a fixed budget, but can instead decide what is the appropriate time to stop. In such systems the used computational budget can be part of the reward function.
(ii) The sequential structure of the current system does not allow simple parallelization, but a PoseAgent that learns to do inference making use of multiple computational cores could be conceived.      

\paragraph*{Acknowledgements.} This work was supported by: European Research Council (ERC) under the European Union's Horizon 2020 research and innovation programme (grant agreement No 647769); The computations were performed on an HPC Cluster at the Center for Information Services and High Performance Computing (ZIH) at TU Dresden.
We thank Joachim Staib for his help with CPU rendering on the Cluster.
\newpage
{\small
\bibliographystyle{ieee}
\bibliography{myLib}
}

\end{document}